\documentclass[a4paper]{cas-sc}

\usepackage[numbers,sort&compress]{natbib}
\def\tsc#1{\csdef{#1}{\textsc{\lowercase{#1}}\xspace}}
\tsc{WGM}
\tsc{QE}
\tsc{EP}
\tsc{PMS}
\tsc{BEC}
\tsc{DE}
\usepackage{float}

\begin{document}
\let\WriteBookmarks\relax
\def\floatpagefraction{1}
\def\textpagefraction{0.001}
\shorttitle{}
\shortauthors{}

%\begin{frontmatter}

\title [mode = title]{Language-Guided Structure-Aware Network for Camouflaged Object Detection}                      
%\tnotemark[1,2]

\author[1]{Min Zhang}
% \cormark[]
\ead{zm321098@163.com}

% \author[1]{Yong Wang}
% %\cormark[1]
% \fnmark[†]
% \ead{ywang@cqut.edu.cn}

% \author[1]{Boran Yang}
% % \cormark[1]
% % \fnmark[†]
% \ead{yangbr@cqut.edu.cn}

% \author[3]{Duoqian Miao}
% %\cormark[1]
% %\fnmark[1]
% \ead{dqmiao@tongji.edu.cn}

\affiliation[1]{organization={School of Artificial Intelligence},
                addressline={Chongqing University of Technology}, 
                city={Chongqing},
                postcode={401120}, 
                country={China}}

% % \affiliation[2]{organization={Goldman Sachs},
% %                 addressline={Orlando}, 
% %                 city={FL},
% %                 postcode={32814}, 
% %                 country={USA}}

% \affiliation[3]{organization={School of Computer Science and Technology},
%                 addressline={Tongji University}, 
%                 city={Shanghai},
%                 postcode={200092}, 
%                 country={China}}

% \cortext[1111]{†Corresponding author}

\begin{abstract}
Camouflaged Object Detection (COD) aims to segment objects that are highly integrated with the background in terms of color, texture, and structure, making it a highly challenging task in computer vision. Although existing methods introduce multi-scale fusion and attention mechanisms to alleviate the above issues, they generally lack the guidance of textual semantic priors, which limits the model’s ability to focus on camouflaged regions in complex scenes. To address this issue, this paper proposes a Language-Guided Structure-Aware Network(LGSAN). Specifically, based on the visual backbone PVT-v2, we introduce CLIP to generate masks from text prompts and RGB images, thereby guiding the multi-scale features extracted by PVT-v2 to focus on potential target regions. On this foundation, we further design a Fourier Edge Enhancement Module (FEEM), which integrates multi-scale features with high-frequency information in the frequency domain to extract edge enhancement features. Furthermore, we propose a Structure-Aware Attention Module (SAAM) to effectively enhance the model’s perception of object structures and boundaries. Finally, we introduce a Coarse-Guided Local Refinement Module (CGLRM) to enhance fine-grained reconstruction and boundary integrity of camouflaged object regions. Extensive experiments demonstrate that our method consistently achieves highly competitive performance across multiple COD datasets, validating its effectiveness and robustness. Our code has been made publicly available. \url{https://github.com/tc-fro/LGSAN}
\end{abstract}


\maketitle

\section{Introduction}
Cod is an important yet challenging task in computer vision, whose core objective is to identify objects that closely resemble their surroundings in terms of color, texture, and shape within complex backgrounds. Due to the inherent lack of saliency in camouflaged objects, they often exhibit blurred boundaries, weakened contours, and extremely low semantic responses, making it difficult for traditional semantic segmentation methods to achieve satisfactory performance on such tasks. 

COD \cite{he2023camouflaged}, \cite{fan2020camouflaged} has significant application value in various fields, such as natural ecological monitoring (e.g., wildlife protection), medical image analysis (e.g., lesion detection), industrial defect inspection, and military concealed target detection. Therefore, developing efficient and robust camouflaged object detection methods holds significant theoretical importance and practical value. Compared with Generic Object Detection (GOD) \cite{pu2024fine}, \cite{pu2023adaptive} and Salient Object Detection (SOD) \cite{zhuge2022salient}, \cite{zhao2019egnet}, the main challenge of COD lies in the high similarity of visual features between the objects and the background, which limits the effectiveness of traditional methods based on saliency or contrast.

In view of the challenge posed by the high similarity between camouflaged objects and the background, recent deep learning–based COD research can be broadly distilled into four concise categories: multi-scale context modeling, bio-inspired mechanism simulation, multi-source information fusion, and multi-task learning. Among them, multi-scale context aims to exploit rich contextual information to capture the diversity of camouflaged objects in appearance and scale, while aggregating cross-layer features and progressively refining representations. Such as HCM \cite{xiao2023concealed}, CamoFormer \cite{yin2024camoformer}, FSPNet \cite{huang2023feature}; Bio-inspired mechanism simulation strategies draw inspiration from the behavioral patterns of natural predators or the human visual detection system. Such as SINet \cite{fan2020camouflaged}, ZoomNet \cite{pang2022zoom}, MFFN \cite{zheng2023mffn}; Multi-source information fusion strategies, in addition to RGB cues, introduce external information such as frequency-domain features, depth, and prompts. Such as FEMNet \cite{zhong2022detecting}, DCE \cite{xiang2021exploring}, CGCOD \cite{zhang2024cgcod}. Multi-task learning aims to jointly optimize multiple related tasks by leveraging shared information and complementary features across tasks. Such as MGL \cite{zhai2021mutual}, FindNet \cite{li2022findnet}, ASBI \cite{zhang2023attention}.

In recent years, the task of COD has still faced many challenges. First, COD models usually rely on a single visual backbone and lack explicit guidance from textual semantic priors, making it difficult to effectively focus on camouflaged regions; Second, camouflaged objects generally suffer from weak boundary information and non-salient structural cues, which poses challenges for precise edge localization and detail restoration; and third, the internal structures of camouflaged regions are complex yet subtle, making it difficult to perform fine-grained modeling of local areas, which in turn affects the regional consistency and structural integrity of segmentation results.

Considering that the detection targets in this study are all known species and that the target categories are available as prior knowledge, text prompts can be introduced to provide category-related semantic information and thus offer the model a clearer focus of attention. Therefore, to address the above issues, this paper proposes a language semantics-guided structure-aware network for camouflaged object detection. First, we input text prompts and RGB images into CLIP to obtain the textual and visual features of camouflaged objects, and employ a text-guided decoder to generate object masks, which guide the multi-scale features output by the PVT-v2 backbone \cite{wang2022pvt} to focus on potential target regions. Subsequently, we design the FEEM, which integrates multi-scale semantic features and extracts high-frequency information in the frequency domain for edge enhancement. The resulting edge enhancement features provide explicit boundary cues for the subsequent modules. On this basis, we propose the SAAM to enhance the model’s perception of object structures and boundaries. However, relying solely on this module still makes it difficult to ensure the coherence and local consistency of target regions. To this end, we further propose the CGLRM, which adopts a dual-branch structure: one branch incorporates channel attention and spatial attention to perform global perception modeling and generate global attention guidance; the other branch divides the input features into four local regions and performs local structural refinement under the guidance of global attention, thereby effectively improving the boundary integrity and structural consistency of camouflaged object segmentation results. 

In summary, our contributions are as follows:
\begin{itemize}
    \item We propose the LGSAN for Camouflaged Object Segmentation, which introduces CLIP to generate region-guided masks, effectively improving the model’s focusing ability on target regions.
    \item We propose the FEEM, which integrates multi-scale semantic information with a frequency-domain modeling strategy to generate edge enhancement features, providing boundary cues for subsequent modules.
    \item We design the SAAM, which integrates structural features of camouflaged objects with edge features to enhance the model’s perception of object structures and boundaries.
    \item We propose the CGLRM, which enhances the structural consistency of target regions by combining spatial partitioning, global guidance fusion, and local refinement.
\end{itemize}

\section{Related Work}
COD initially originated from traditional image-level methods, which relied on manually designed low-level features (such as texture, intensity, and color) to capture subtle differences between the foreground and background, thereby laying the foundation for early research in this field. Although traditional methods achieve certain effectiveness in low-complexity scenarios, they often fail in cases of low resolution or high similarity between foreground and background, and their feature representation capability is limited. With the development of deep learning, end-to-end approaches for learning complex representations have gradually become mainstream.

Recent deep learning-based COD methods can be broadly categorized into four representative strategies: multi-scale context modeling, bio-inspired mechanism simulation, multi-source information fusion, and multi-task learning. Among them, multi-scale context aims to exploit rich contextual information to capture the diversity of camouflaged objects in appearance and scale, while aggregating cross-layer features and progressively refining representations. For example, HCM \cite{xiao2023concealed} focuses on low-confidence regions through a reversible recalibration mechanism, thereby detecting parts that are initially overlooked; CamoFormer \cite{yin2024camoformer} introduces masked separable attention to achieve top-down multi-level feature refinement; FSPNet \cite{huang2023feature} designs a non-local token enhancement mechanism to improve feature interaction capability, and incorporates a feature shrinkage decoder to optimize the results; OWinCANet \cite{li2023cross} introduces cross-layer overlapping window attention based on a shifted window strategy, achieving a balance between local and global information through sliding-aligned windows.

Bio-inspired mechanism simulation strategies draw inspiration from the behavioral patterns of natural predators or the human visual detection system, typically adopting a multi-stage, coarse-to-fine process to progressively improve the localization and segmentation accuracy of camouflaged objects. SINet \cite{fan2020camouflaged} is inspired by the first two stages of hunting and consists of two main modules: the Search Module and the Identification Module. The former is responsible for searching for camouflaged objects, while the latter is used to precisely detect them; ZoomNet \cite{pang2022zoom}, which mimics how humans observe vague images by zooming in and out, employs this zoom strategy—together with a designed scale integration unit and a hierarchical mixed-scale unit—to learn discriminative mixed-scale semantics; MFFN \cite{zheng2023mffn} acquires complementary information from multiple views (different angles, distances, and perspectives), thereby effectively handling complex scenarios involving camouflaged objects.

Multi-source information fusion strategies, in addition to RGB cues, introduce external information such as frequency-domain features, depth, and prompts to enhance the discriminative power and robustness of COD. Frequency-domain methods (e.g., FEMNet \cite{zhong2022detecting}) use frequency-domain information as a supplementary cue to improve the detection of camouflaged objects. Depth-based methods, such as DCE \cite{xiang2021exploring}, introduce auxiliary depth estimation and a GAN-based multi-modal confidence loss. Prompt-based methods, such as CGCOD \cite{zhang2024cgcod}, combine visual and textual prompts to enhance the perception of camouflaged scenes. Multi-source information fusion strategies, in addition to RGB cues, introduce external information such as frequency-domain features, depth, and prompts.

Multi-task learning in camouflaged object detection aims to jointly optimize multiple related tasks, leveraging shared information and complementary features across tasks to significantly enhance the model’s discriminative capability and generalization performance. Within this framework, boundary-supervised methods (e.g., MGL \cite{zhai2021mutual}, FindNet \cite{li2022findnet}, ASBI \cite{zhang2023attention}) introduce edge-detection branches, enabling the model to better capture the edge details of camouflaged objects, thereby effectively improving boundary accuracy and the representation of object details.

\begin{figure}[!h]
\centering
\includegraphics[width=\textwidth]{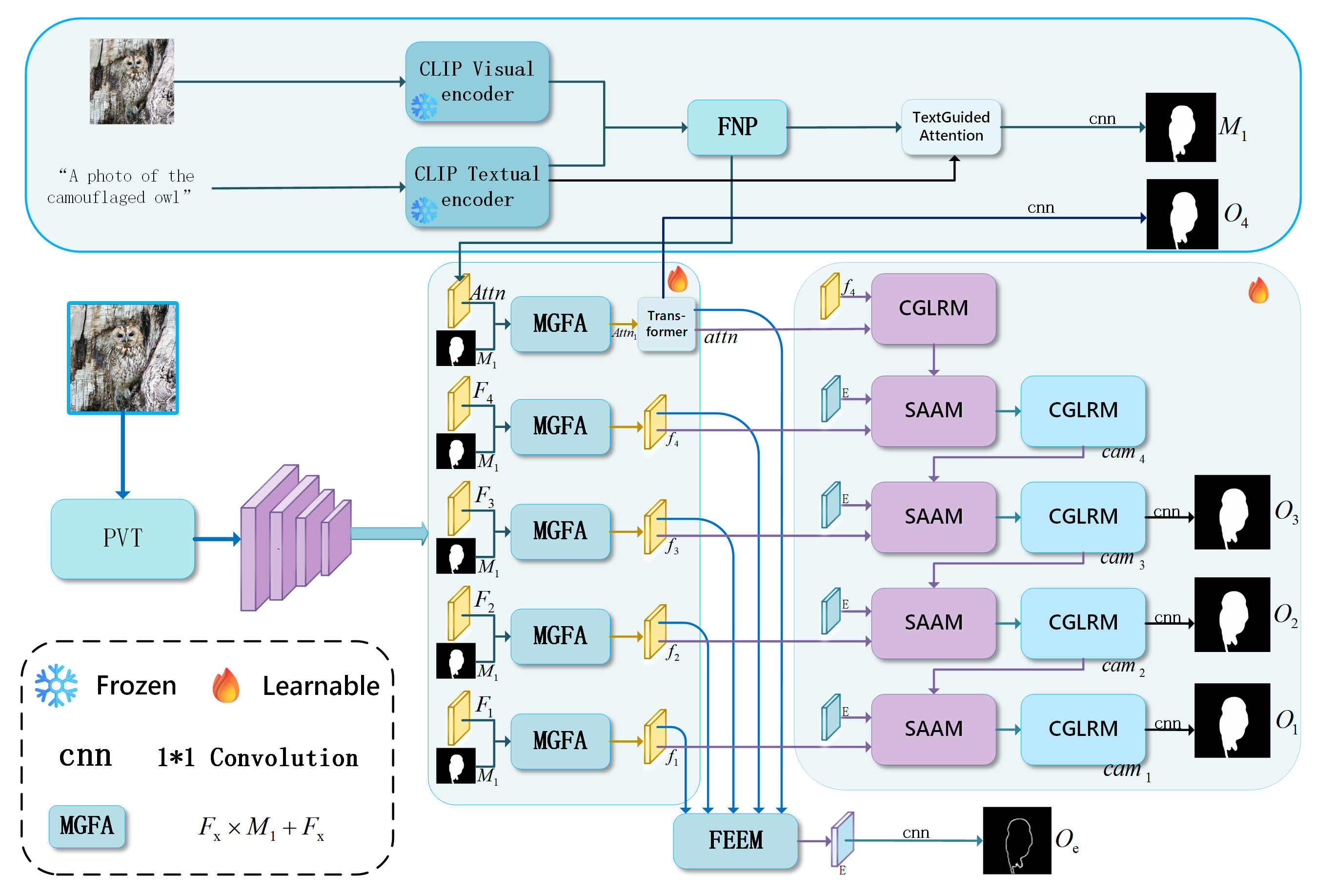}
\caption{The architecture of LGSAN. The overall framework of the model consists of five key components: the PVT-v2 backbone, the CLIP backbone, the FEEM, the SAAM, and the CGRLM. Refer to Section 3 for details.}
\label{network}
\end{figure}

\section{METHODOLOGY}
\subsection{Network Overview}
As shown in Fig.~\ref{network}, we propose a Language-Guided Structure-Aware Network (LGSAN) for Camouflaged Object Detection. The overall framework of the model consists of five key components: the PVT-v2-b3 backbone, the CLIP backbone, the FEEM, the SAAM, and the CGRLM. Refer to Section 3 for details.

For the visual backbone, we adopt PVT-v2-b3 as the image feature extractor to obtain four-scale feature maps from the input image $I \in \mathbb{R}^{H \times W \times 3}$, denoted as ${F_i}_{i=1}^{4}$. To enhance the semantic information of camouflaged objects, we further introduce a frozen CLIP model: the text encoder extracts textual features from text prompts (e.g., “a photo of the camouflaged owl”), while the visual encoder extracts multi-scale visual features from its 8th, 16th, and 24th layers. Subsequently, cross-scale alignment and fusion are performed via a FPN \cite{wang2022cris} to obtain the CLIP visual features $Attn$, which are combined with the textual features and fed into the Text-Guided decoder to generate the object mask $M_1$. The mask is applied to $\{F_i\}_{i=1}^4$ and $Attn$ through the MGFA operation, as shown in Fig.~\ref{network}, to achieve explicit feature enhancement, resulting in $\{F_i\}_{i=1}^4$ and $Attn_1$, which enable the model to focus on potential camouflaged regions.

The enhanced features $Attn_1$ are first fed into a transformer to obtain the $attn$ features. Subsequently, the Fourier Edge Enhancement Module (FEEM) takes $\{F_i\}_{i=1}^4$ and $attn$ as inputs to generate the edge enhancement features $E$.

In the next stage, the highest-level feature $f_4$ is concatenated with $attn$ and passed through the CGRLM to obtain coarse-grained structural features.
These features, together with $E$ and $attn$, are then input into the SAAM to produce structure-enhanced features, which are subsequently refined by the CGRLM to generate $cam_4$. For the remaining scales $i = 3, 2, 1$, the refinement process begins by upsampling the previous output $cam_{i+1}$ and concatenating it with the corresponding backbone feature $f_i$. A convolution block is applied to compress the result into $\bar f_i$, which, along with the edge guidance $E$ and $cam_{i+1}$, is fed into the SAAM. Finally, local refinement is performed via the CGRLM, yielding the output $cam_i$.

In the prediction stage, the network outputs segmentation results at multiple scales through convolution: intermediate prediction maps $O_3, O_2, O_1$ are obtained from $cam_3, cam_2, cam_1$, respectively, and are upsampled to the original resolution via interpolation; $O_{4}$ is generated from $attn$ as the semantic-guided prediction, $O_e$ is derived from $E$ as the edge prediction, and the mask $M_1$ is also included. The final output set is
\[
\mathcal{O} = \{O_1, O_2, O_3, O_{4}, M_{1}, O_e\}.
\]

\begin{figure}[!htbp]
\centering
\includegraphics[width=\textwidth]{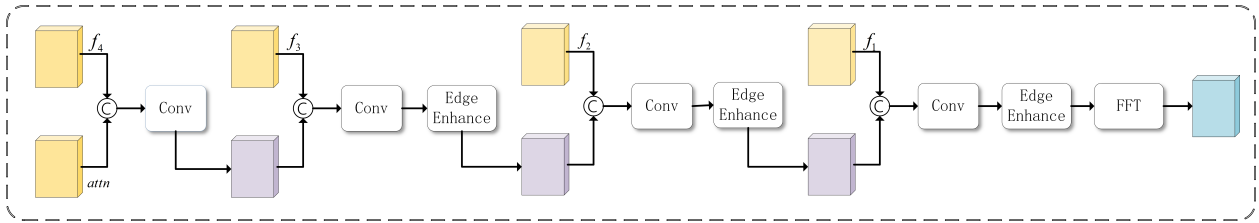}
\caption{The architecture of the FEEM. The FEEM generates edge enhancement features through multi-scale fusion, edge enhancement, and high-frequency modeling in the frequency domain.}
\label{FEEM}
\end{figure}

\subsection{Fourier Edge Enhancement Module}
As shown in Fig.~\ref{FEEM}, the FEEM takes the features $\{F_i\}_{i=1}^4$ and $attn$ as inputs. First, the highest-level feature $f_4$ is concatenated with $attn$ and passed through a channel compression convolution to obtain the high-level fused features:
\begin{align}
f_4' = \mathrm{Conv}\big(\mathrm{C}(\mathrm{Up}(f_4), attn)\big).
\end{align}

The feature is then progressively upsampled and concatenated with the lower-level features, followed by convolutional fusion and enhancement of boundary representations through the EdgeEnhancer \cite{gao2024multi} module:
\begin{align}
f_3' = \mathrm{Enh}\big(\mathrm{Conv}(\mathrm{C}(\mathrm{Up}(f_4'), f_3))\big),
\end{align}
\begin{align}
f_2' = \mathrm{Enh}\big(\mathrm{Conv}(\mathrm{C}(\mathrm{Up}(f_3'), f_2))\big),
\end{align}
\begin{align}
f_1' = \mathrm{Enh}\big(\mathrm{Conv}(\mathrm{C}(\mathrm{Up}(f_2'),f_1))\big).
\end{align}
Here, $\mathrm{C}$ denotes feature concatenation, $\mathrm{Up}$ denotes feature upsampling, $\mathrm{Enh}$ denotes the \textit{EdgeEnhancer} module, 
and $\mathrm{Conv}$ denotes convolution.

The EdgeEnhancer is based on the idea of average pooling difference, where input features are smoothed and their differences are computed to extract high-gradient regions (i.e., edges). An edge weight map is then generated and fused with the original features in a residual manner, thereby explicitly enhancing boundary contrast in the spatial domain. Its mathematical formulation can be expressed as:
\begin{align}
E_{\mathrm{diff}} = X - \mathrm{Pool}(X),
\end{align}
\begin{align}
W_{\mathrm{edge}} = \sigma\big(\mathrm{BN}(\mathrm{Conv}_{1\times1}(E_{\mathrm{diff}}))\big),
\end{align}
\begin{align}
X_{\mathrm{enh}} = X + W_{\mathrm{edge}}.
\end{align}
Here, $\mathrm{Pool}(\cdot)$ denotes a $3 \times 3$ average pooling operation, $\sigma(\cdot)$ represents the Sigmoid function, and $W_{\mathrm{edge}}$ indicates the edge weight map.

To further extract high-frequency detail information of object boundaries, we apply a Fourier Transform on the fused low-level feature $f_1'$, explicitly capturing high-frequency responses:
\begin{align}
E_{\text{high}} = \text{ReLU}\big(\text{FFT}_{\text{high}}(X_1')\big).
\end{align}
Finally, edge enhancement features are output:
\begin{align}
E = \text{Reshape}(e_{\text{high}}).
\end{align}

\begin{figure}[!htbp]
\centering
\includegraphics[width=\textwidth]{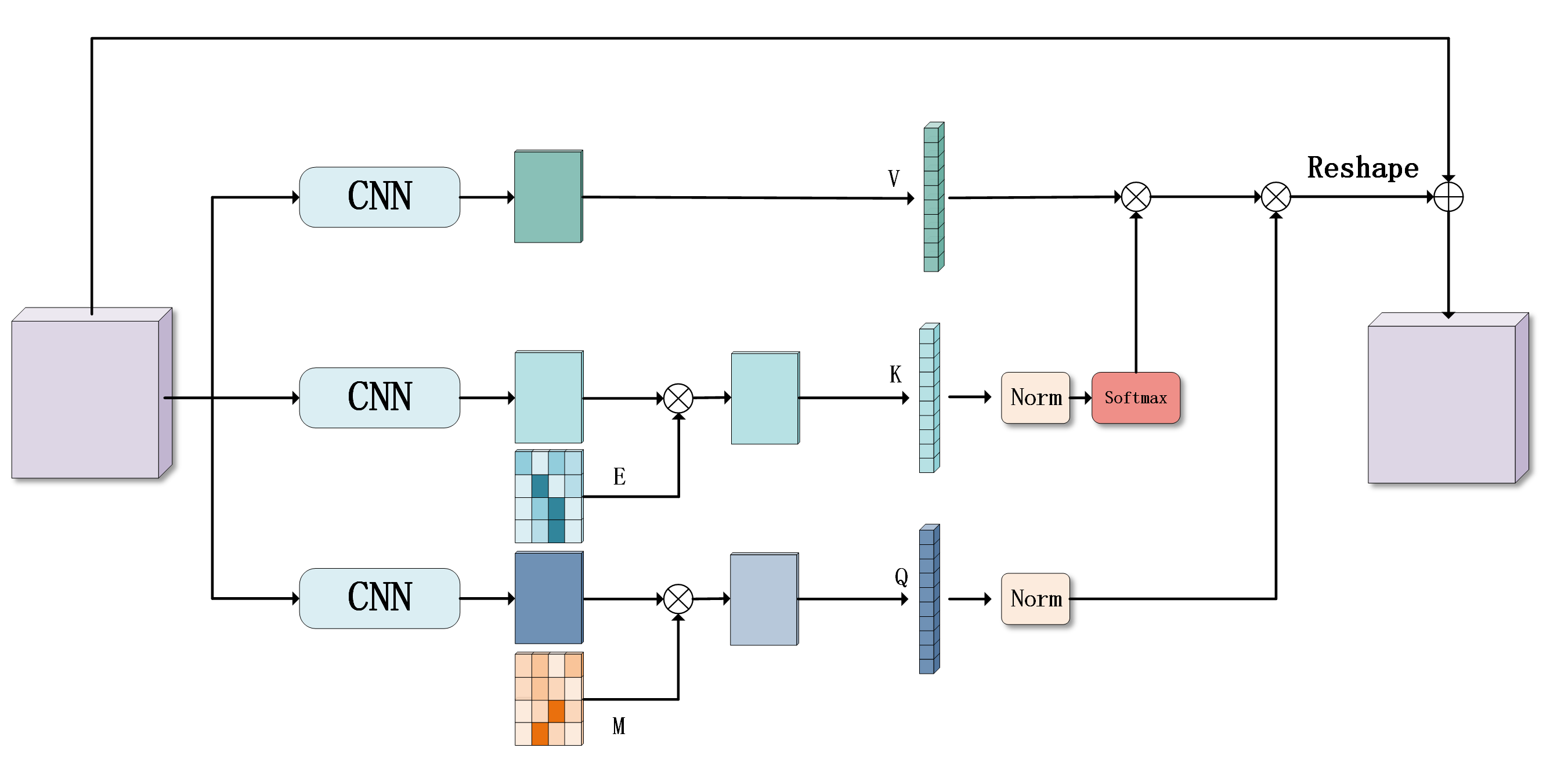}
\caption{The SAAM introduces semantic information of camouflaged objects into a lightweight attention framework to highlight camouflaged regions, while incorporating edge enhancement features to emphasize boundary information, thereby enabling the model to focus on structural and boundary details of camouflaged objects at high resolution.}
\label{SAAM}
\end{figure}

\subsection{Structure-Aware Attention Module}
As shown in Fig.~\ref{SAAM}, to enhance the model’s ability to discriminate boundary details and object structures of camouflaged targets, we propose the SAAM. Unlike traditional self-attention mechanisms that equally consider all spatial positions, the SAAM incorporates the semantic information $M$ of camouflaged objects (i.e., $cam_{i}$ in Fig.~\ref{network}, which is rich in semantic cues of camouflaged targets) and the edge-guided features $E$ (i.e., $E$ in Fig.~\ref{network}, which contains boundary information) to achieve task-specific guided attention modeling.
Specifically, given an input feature map $x \in \mathbb{R}^{B \times C \times H \times W}$ (i.e., the \(\bar{f}_i\) in Section~3.1), we first apply convolutions to extract the query (Q), key (K), and value (V) features, thereby reducing computational cost. Then, we incorporate external guidance into the attention modeling: the semantic information $M$ of camouflaged objects is applied to the query vectors (emphasizing the structure of camouflaged targets), while the edge enhancement features $E$ are applied to the key vectors (highlighting boundary information), yielding:

\begin{align}
Q &= \operatorname{Norm}\!\big(\mathrm{CNN}(x) \otimes M\big),
\end{align}
\begin{align}
K &= \operatorname{Norm}\!\big(\mathrm{CNN}(x) \otimes E\big),
\end{align}
\begin{align}
V &= \text{CNN}(x).
\end{align}
Here, $\mathrm{CNN}$ denotes convolution, $\otimes$ denotes element-wise multiplication, and $\mathrm{Norm}$ denotes vector normalization. Subsequently, $Q$, $K$, and $V$ are reshaped from $[B,C,H,W]$ to the token representation $[B,HW,C]$. To further reduce memory consumption, we adopt an approximate attention computation strategy: the transposed key features are first processed with softmax, then multiplied by the value vectors, and finally multiplied by the query vectors. The computation process is as follows:

\begin{align}
A &= \text{Softmax}(K^\top),
\end{align}
\begin{align}
AV &= A \cdot V,
\end{align}
\begin{align}
\text{Output} &= Q \cdot AV.
\end{align}
Here, $K^\top$ denotes the transpose over the last two axes, $\cdot$ denotes batched matrix multiplication, in Fig.~\ref{SAAM}, $\oplus$ denotes element-wise addition. The output features are reshaped to restore the spatial structure and fused with the original input features through residual connection.

\begin{figure}[htbp]
\centering
\includegraphics[width=\textwidth]{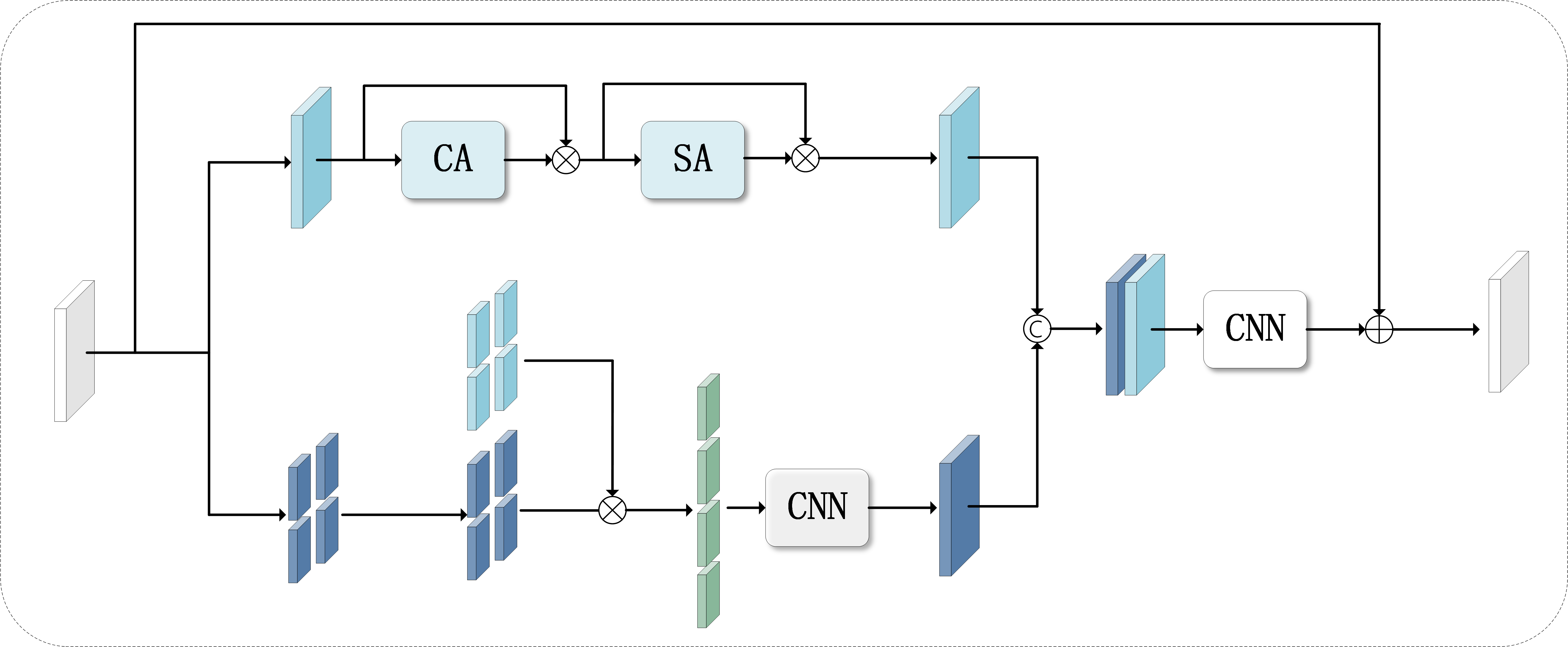}
\caption{The CGLRM employs channel and spatial attention to obtain global guidance, performs local refinement through 2×2 spatial partitioning, thereby ensuring structural consistency and boundary integrity.}
\label{CGLRM}
\end{figure}
\subsection{Coarse-Guided Local Refinement Module}
As shown in Fig.~\ref{CGLRM}, we propose the CGLRM. Given an input feature map $x \in \mathbb{R}^{B \times C \times H \times W}$, we first apply Channel Attention (CA) and Spatial Attention (SA) to extract attention-enhanced features:
\begin{align}
x_{\text{ca}} = \text{CA}(x) \otimes x,
\end{align}
\begin{align}
g = \text{SA}(x_{\text{ca}}) \otimes x_{\text{ca}}.
\end{align}
Here, $\otimes$ denotes element-wise multiplication. The attention-enhanced features $g$ provide coarse-grained spatial guidance for the target.
Next, we spatially divide the input feature $x$ into four non-overlapping sub-regions:
\begin{equation}
\{x_i\}_{i=1}^{4} = \text{SpatialSplit}(x).
\end{equation}

The attention-enhanced features $g$ are divided in the same manner into $\{g_i\}_{i=1}^{4}$, ensuring that each local region is equipped with the corresponding global guidance information. This spatial partitioning strategy effectively suppresses irrelevant interference by introducing global guidance into the corresponding regions; meanwhile, it maintains the independence of each region while enhancing overall structural coherence and the ability to restore fine details. As follows:
\begin{equation}
\hat{x}_i = \text{LocalRefine}\big(x_i \otimes \sigma(g_i)\big), \quad i \in \{1,2,3,4\}.
\end{equation}
Here, $\sigma()$ denotes the Sigmoid activation used to generate smooth guidance weights, and LocalRefine denotes a convolution operation. All local refinement results are concatenated into a complete feature map:
\begin{equation}
x_{\text{local}} = \text{C}(\{\hat{x}_i\}_{i=1}^4).
\end{equation}
Subsequently, it is fused with the attention-enhanced features:
\begin{equation}
x_{\text{fuse}} = \text{CNN}\big(\text{C}(x_{\text{local}}, g)\big),
\end{equation}
where \(\text{CNN}\) denotes convolution, and \(\text{C}\) denotes feature concatenation. And finally passed through the Output Enhancement Module to generate the final output:
\begin{equation}
\text{Output} = \text{CNN}(x_{\text{fuse}}) \oplus x,
\end{equation}
where \(\text{CNN}\) denotes convolution, \(\oplus\) denotes element-wise addition, and \(\text{C}\) denotes feature concatenation. This design explicitly models local structures under global guidance, effectively improving boundary integrity and structural consistency.

\subsection{Loss Function}
In this model, the loss function from the boundary-guided camouflaged object detection network (BGNet) proposed by Sun et al. \cite{sun2022boundary} is adopted to improve COD performance. The total loss function consists of two parts: the camouflaged object mask $(G_o)$ and the camouflaged object boundary $(G_e)$. For the camouflaged object mask, the weighted binary cross-entropy loss $({L^{W}}{BCE})$ and the weighted IoU loss $({L^{W}}{IOU})$ are combined, following the approach of Wei et al. \cite{wei2020f3net}. For boundary prediction, the Dice loss $L_{\mathrm{dice}}$ \cite{xie2020segmenting} is used as the boundary supervision signal. Therefore, the total loss function is defined as follows:
\begin{multline}
L_{\text{total}} =
\sum_{i=1}^{4} \Big( L^{\mathrm{W}}_{\mathrm{BCE}}(O_i, G_o)
+ L^{\mathrm{W}}_{\mathrm{IoU}}(O_i, G_o) \Big) 
+ L^{\mathrm{W}}_{\mathrm{BCE}}(M_1, G_o)
+ L^{\mathrm{W}}_{\mathrm{IoU}}(M_1, G_o)
+ \lambda L_{\mathrm{dice}}(O_e, G_e),
\end{multline}
where $\lambda$ is a weighting parameter used to balance mask supervision and edge supervision. In the experiments, $\lambda$ is set to 5.

\section{EXPERIMENT}
\subsection{Implementation Details}
In this study, the model is implemented based on the PyTorch framework and trained and tested on an NVIDIA GeForce RTX 5000 GPU. During training, all input images are uniformly resized to a resolution of 521×521. The Adam optimizer \cite{kingma2014adam} is employed, with the training process set to 25 epochs and a batch size of 4. The initial learning rate is set to 0.0001, and a poly learning rate decay strategy is adopted with the decay power parameter set to 0.9 to dynamically adjust the learning rate. During testing, the input images are also resized to 521×521, and after inference, the outputs are restored to their original resolution for model performance evaluation.

\subsection{Datasets}
To assess the effectiveness of the proposed approach, we evaluate it on three standard camouflaged object detection benchmarks—CAMO \cite{le2019anabranch}, COD10K \cite{fan2020camouflaged}, and NC4K \cite{lv2021simultaneously}. LGSAN is trained on the training splits of CAMO and COD10K, and evaluated on their official test sets as well as the held-out NC4K test set, providing a comprehensive assessment of both detection performance and generalization.

\subsection{Evaluation Metrics}
In image-level camouflaged object detection tasks, the commonly used evaluation metrics mainly include Mean Absolute Error (MAE, M) \cite{perazzi2012saliency}, Weighted F-measure ($F^w_\beta$) \cite{margolin2014evaluate}, Structural Similarity Index ($S_\alpha$) \cite{fan2017structure}, and mean E-measure ($E_\phi$) \cite{fan2021cognitive}.

\begin{figure*}[htbp]
\centering
\includegraphics[width=\textwidth]{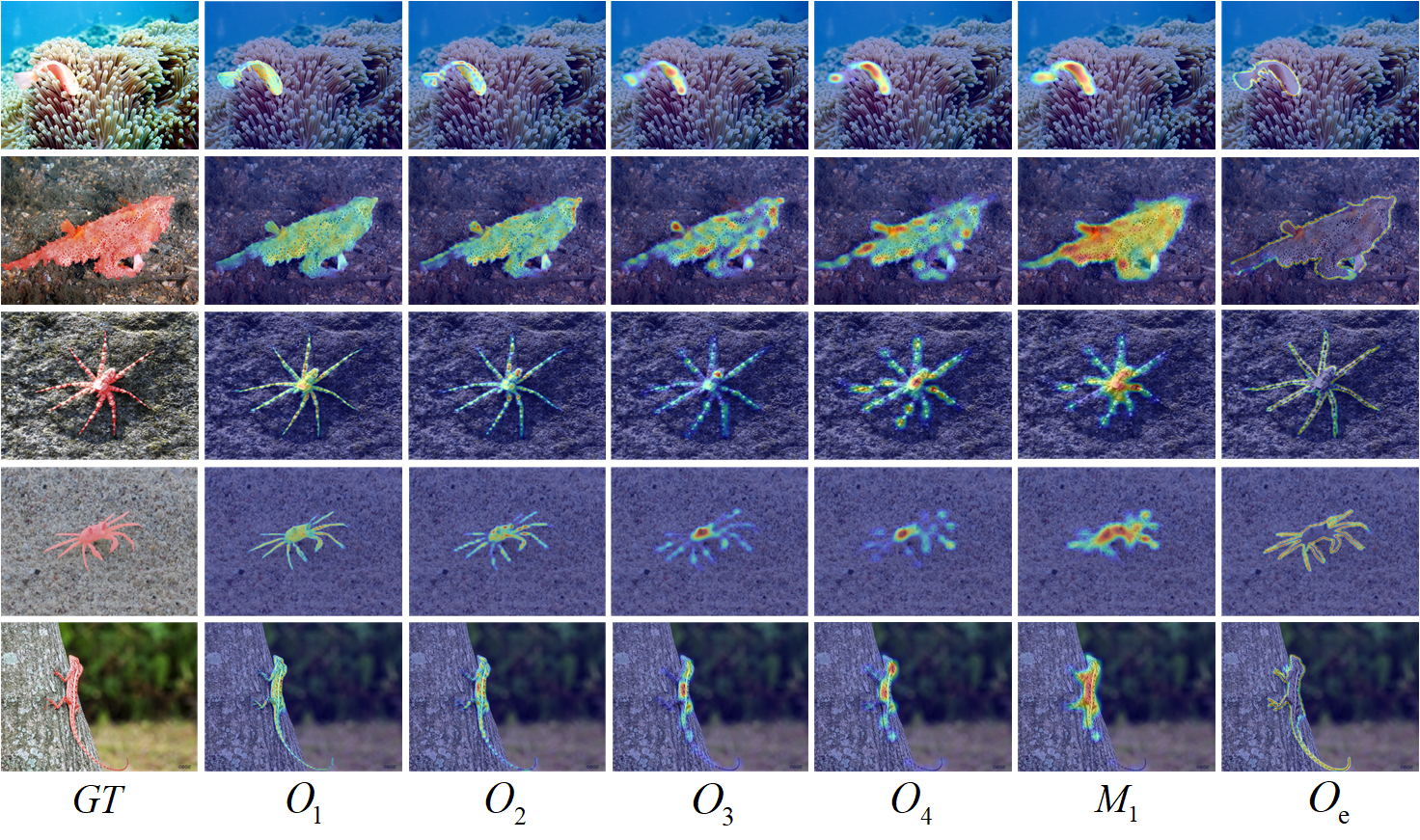}
\caption{The heatmaps of LGSAN from $O_4$ to $O_1$ illustrate the progressive refinement process, while the heatmaps of $M_1$ and $O_e$ are also presented for comparison.}
\label{HEAT}
\end{figure*}

\subsection{Compared With the State-of-the-Art Methods}
To thoroughly assess the effectiveness of the proposed LGSAN, we perform extensive comparisons with representative recent methods, including BGNet \cite{sun2022boundary}, ZoomNet \cite{pang2022zoom}, EVP \cite{liu2023explicit}, EAMNet \cite{sun2023edge}, FSPNet \cite{huang2023feature}, FEDER \cite{he2023camouflaged}, DCNet \cite{yue2023dual}, SARNet \cite{xing2023go}, DINet \cite{zhou2024decoupling}, SDRBet \cite{guan2024sdrnet}, VSCode \cite{luo2024vscode}, FSEL \cite{sun2024frequency}, CamoFormer \cite{yin2024camoformer}, IPNet\cite{Ipnet}, CODdiff \cite{CODdiff}, ESNet \cite{bi2025edge}, SENet \cite{hao2025simple}, KCNet \cite{wu2025knowledge}, UGDNett \cite{yang2025uncertainty}, BDCLNet \cite{zhao2025bilateral}, and CGCOD \cite{zhang2024cgcod}, on three mainstream COD benchmarks (CAMO, COD10K, and NC4K).

\begin{table*}[ht]
    \centering
    \caption{Quantitative comparison with state-of-the-art methods for COD on 3 benchmarks using 4 widely used evaluation metrics ($S_\alpha, E_{\phi}, F^w_\beta, M$). “↑” / “↓” indicates that larger/smaller is better. The best results are in $\textbf{bold}$,**** means the model was not tested on this dataset.}
    \label{tab1}
    \begin{tabular}{r|c|p{0.65cm}p{0.65cm}p{0.65cm}p{0.65cm}|p{0.65cm}p{0.65cm}p{0.65cm}p{0.65cm}|p{0.65cm}p{0.65cm}p{0.65cm}p{0.65cm}}
        \hline
        \multirow{2}{*} {Method} & \multirow{2}{*} {Pub./Year} & \multicolumn{4}{|c|}{CAMO} & \multicolumn{4}{|c|}{COD10K} & \multicolumn{4}{|c}{NC4k} \\
        \cline{3-14}
         & & $S_\alpha \uparrow$ & $E_{\phi} \uparrow$ & $F^w_\beta \uparrow$ & $M \downarrow$ & $S_\alpha \uparrow$ & $E_{\phi} \uparrow$ & $F^w_\beta \uparrow$ & $M \downarrow$ & $S_\alpha \uparrow$ & $E_{\phi} \uparrow$ & $F^w_\beta \uparrow$ & $M \downarrow$ \\
        \hline
        BGNet & IJCAI'22 & 0.812 & 0.870 & 0.749 & 0.073 & 0.831 & 0.901 & 0.722 & 0.033 & 0.851 & 0.907 & 0.788 & 0.044 \\
        ZoomNet & CVPR'22 & 0.820 & 0.892 & 0.752 & 0.066 & 0.838 & 0.911 & 0.729&0.029 & 0.853 & 0.912 & 0.784 & 0.059 \\
        EVP & CVPR'23  & {0.846} & {0.895} & {0.777} & {0.059} & {0.843} & {0.907} & {0.742} & {0.029} & **** & **** & **** & **** \\
        EAMNet & ICME'23  & 0.831 & 0.890 & 0.763 & 0.064 & 0.839 & 0.907 & 0.733 & 0.029 & {0.862} & {0.916} & {0.801} & {0.040} \\
        FSPNet & CVPR'23  & {0.856} & {0.899} & {0.799} &{0.050} & {0.851} & {0.895} & {0.735} & {0.026} &  {0.879} & {0.915} & {0.816} & {0.035} \\
        FEDER & CVPR'23  & {0.822} & {0.886} & {0.809} & {0.067} & {0.851} & {0.917} & {0.752} & {0.028} &  {0.863} & {0.917} & {0.827} & {0.042} \\
        DCNet & TCSVT'23  & {0.870} & {0.922} & {0.831} & {0.050} & {0.873} & {0.934} & {0.810} & {0.022} & **** & **** & **** & **** \\
        SARNet & TCSVT'23 & {0.868} &{0.927} & {0.828} &  {0.047} & {0.864} &  {0.931} &  {0.777} &  {0.024} &  {0.886} & \textbf{0.937} &  {0.842} & {0.032} \\
        DINet & TMM'24 & {0.821} & {0.874} & {0.790} &  {0.068} & {0.832} &  {0.903} &  {0.761} &  {0.031} &  {0.856} & {0.909} &  {0.824} & {0.043} \\
        SDRNet & KBS'24 & {0.872} & {0.924} & {0.826} &  {0.049} & {0.871} &  {0.924} &  {0.785} &  {0.023} &  {0.889} & {0.934} &  {0.842} & {0.032} \\
        % CamoFocus & WACV'24 & {0.873} & {0.926} & {0.842} &  {0.043} & {0.873} &  {0.935} &  {0.802} &  {0.021} &  {0.889} & {0.936} &  {0.853} & {0.030} \\
        VSCode & CVPR'24 & {0.873} & {0.925} & {0.844} &  {0.046} & {0.869} &  {0.931} &  {0.806} &  {0.023} &  {0.891} & {0.935} &  {0.863} & {0.032} \\
        FSEL & ECCV'24 & {0.885} & {0.942} & {0.857} &  {0.040} & {0.877} &  {0.937} &  {0.799} &  {0.021} &  {0.892} & {0.941} &  {0.852} & {0.030} \\
        % CenSAM & AAAI'24 & {0.719} & {0.775} & {0.659} &  {0.113} & {0.775} &  {0.838} &  {0.681} &  {0.067} & **** & **** & **** & **** \\
        CamoFormer & TPAMI'24 & {0.876} & {0.930} & {0.856} & {0.043} & {0.838} &  {0.916} &  {0.753} &  {0.029} &  {0.888} & {0.937} &  {0.863} & {0.031} \\
        IPNet & EAAI'24 &0.864 & 0.924 & 0.836 & 0.047 & 0.850 & 0.922 & 0.785 & 0.026 & **** & **** & **** & **** \\
        CODdiff & KBS'25 & 0.839 & 0.911 & 0.802 & 0.054 & 0.837 & 0.919 & 0.759 & 0.026 & 0.865 & 0.926 & 0.827 & 0.036 \\
        ESNet & IVC'25 & {0.860} & {0.918} & {0.846} & {0.050} & {0.864} &  {0.933} &  {0.803} &  {0.024} &  {0.880} & {0.931} &  {0.856} & {0.035} \\
        SENet & TIP'25 & {0.888} & {0.932} & {0.847} & {0.039} & {0.865} &  {0.925} &  {0.780} &  {0.024} &  {0.889} & {0.933} &  {0.843} & {0.032} \\
        KCNet & EAAI'25 & {0.882} & {0.934} & {0.847} & {0.039} & {0.865} &  {0.925} &  {0.780} &  {0.024} &  {0.889} & {0.933} &  {0.843} & {0.032} \\
        UGDNet & TMM'25 & {0.888} & {0.942} & {0.865} & {0.038} & {0.885} &  {0.947} &  {0.822} &  {0.019} &  {0.895} & {0.943} &  {0.862} & {0.028} \\
        BDCLNet & KBS'25 & {0.881} & {0.929} & {0.845} & {0.039} & {0.869} &  {0.935} &  {0.790} &  {0.022} &  {0.888} & {0.932} &  {0.844} & {0.032} \\
        CGCOD & ACMMM'25 & \textbf{0.896} & {0.947} & {0.864} & {0.036} & {0.890} &  {0.948} &  {0.824} &  \textbf{0.018} &  \textbf{0.904} &\textbf {0.949} &  \textbf{0.869} & \textbf{0.026} \\
        LGSAN &Ours & \textbf{0.896} & \textbf{0.949} & \textbf{0.870} & \textbf{0.034} & \textbf{0.894} & \textbf{0.950} & \textbf{0.833} & \textbf{0.018} & {0.903} & \textbf{0.949} & \textbf{0.869} & \textbf{0.026} \\
  
        %\textcolor{red}{2222}
        \hline
    \end{tabular}
\end{table*}

% \begin{figure}[h] % figure* 改为 figure
% \centering
% \includegraphics[width=10cm]{hot_att.png} % width从textwidth改为linewidth
% \caption{$E\_att$ represents edge attention, and $F_1$ to $F_4$ are the target feature maps after frequency domain enhancement, guided by edge attention. Specifically, $E\_att$ focuses on the image edges, helping the model effectively differentiate between the foreground and background. Based on this guidance, $F_1$ to $F_4$ show the enhanced feature maps at different scales, significantly improving the target's detail and global structure, particularly in terms of texture and shape. By combining edge guidance and frequency domain enhancement, the model is better able to capture the features of camouflaged objects, thereby improving detection accuracy.}
% \label{hot}
% \end{figure}
\begin{figure}[htbp]
\centering
\includegraphics[width=\textwidth]{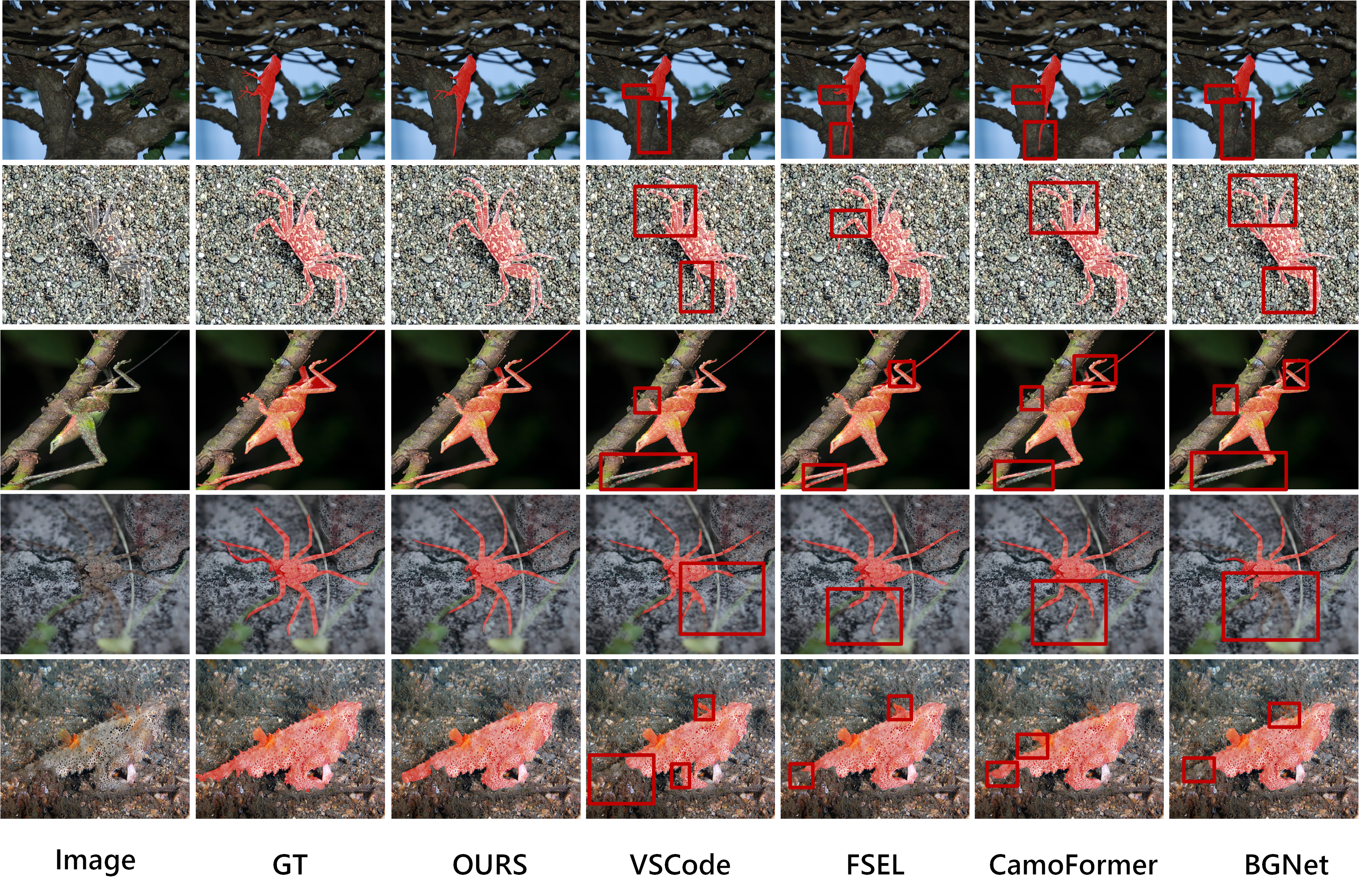}
\caption{Compared with existing methods, the proposed approach achieves superior performance in terms of localization accuracy, structural integrity, and boundary details across multiple categories of camouflaged objects.}
\label{compare_sota}
\end{figure}

\subsubsection{Quantitative Evaluation}

Table~\ref{tab1} presents the quantitative comparison between LGSAN and existing methods in terms of four commonly used metrics: $S_\alpha$, $E_{\phi}$, $F^w_\beta$, and $M$. It can be observed that LGSAN achieves consistently competitive performance across almost all metrics on the three datasets.
(1) On CAMO, LGSAN ranks first on $E_{\phi}$, $F^w_\beta$, and $M$, and achieves a score of 0.896 on $S_\alpha$, tying with CGCOD for the best result.
(2) On COD10K, LGSAN achieves the best performance across all four metrics, with $S_\alpha$ = 0.894, $F^w_\beta$ = 0.833, and $M$ = 0.018.
(3) On NC4K, LGSAN ties with CGCOD for the best results on $E_{\phi}$, $F^w_\beta$, and $M$, while showing only a marginal gap of 0.001 on $S_\alpha$.
Overall, LGSAN not only maintains structural consistency but also significantly enhances boundary details, demonstrating strong generalization and robustness.

\subsubsection{Qualitative Evaluation}
To provide a more intuitive validation of the segmentation performance of the proposed LGSAN, qualitative visual analyses are conducted from two perspectives:

(1) Evolution of outputs across decoding stages.
Fig.~\ref{HEAT} illustrates the predictions and attention distributions of LGSAN at different decoding stages. Specifically, $O_1$–$O_4$ denote the outputs from the four decoding stages, $M_1$ is the semantic mask generated by CLIP, and $O_e$ is the edge prediction from FEEM. As shown, the shallow outputs ($O_4$, $O_3$) can roughly localize targets but still suffer from blurry boundaries. With progressive decoding ($O_2$, $O_1$), the network gradually suppresses background noise and refines object contours, enabling a transition from coarse localization to structurally detailed, high-quality predictions. In this process, $M_1$ provides stable global attention, while $O_e$ delivers explicit boundary cues; their synergy enhances the structural consistency and boundary integrity of the final results.

(2) Comparison with representative SOTA methods.
Fig.~\ref{compare_sota} presents comparisons between LGSAN and representative methods, including VSCode, FSEL, CamoFormer, and BGNet, under diverse challenging scenarios. It can be observed that LGSAN consistently produces masks with clear boundaries, coherent structures, and rich details across various target categories (e.g., insects, crustaceans, camouflaged animals). Even in cases where target and background textures are highly similar, LGSAN accurately separates the target regions. In contrast, other methods often suffer from fractures, boundary omissions, or excessive smoothing in fine parts such as limbs and antennae. These visual results demonstrate that LGSAN maintains superior target consistency and detail fidelity under complex backgrounds and low-contrast conditions, yielding predictions closest to the ground truth (GT).

\begin{table}[ht]
    \centering
    \caption{Ablation study of LGSAN. In the table, “B” denotes the baseline composed of PVT-v2-B3 and a simple CNN, “E” represents the FEEM module, “c” indicates the use of CLIP for generating the $M_1$ mask, and “SC” refers to the SAAM and CGLRM.}
    \label{tab2}
    \begin{tabular}{c|l|c|c|c|c}
        \hline
        \multirow{2}{*}{Model} & \multirow{2}{*}{Method} & \multicolumn{4}{c}{COD10K} \\
        \cline{3-6}
        & & $S_\alpha \uparrow$ & $E_{\phi} \uparrow$ & $F^w_\beta \uparrow$ & $M \downarrow$ \\
        \hline
        a & B & 0.831 & 0.905 & 0.706 & 0.032 \\
        b & B + C & 0.890 & 0.947 & 0.825 & 0.019 \\
        c & B + C + E & 0.893 & 0.948 & 0.830 & 0.018 \\
        d & B + C + E + SC & 0.894 & 0.950 & 0.833 & 0.018 \\
        \hline
    \end{tabular}
\end{table}
\subsection{Ablation Study}
To comprehensively validate the contributions of each key module to camouflaged object segmentation, we conducted step-by-step ablation studies on the COD10K dataset (see Table~\ref{tab2}). Starting from the backbone network as the baseline, we progressively introduced CLIP, FEEM, and the jointly designed SAAM+CGLRM module, thereby systematically analyzing the performance improvements and underlying mechanisms of each component.

\subsubsection{Baseline (B)}
The baseline model adopts the PVT-V2-B3 backbone coupled with a simple CNN decoder (from SARNet-H~\cite{xing2023go}), relying solely on visual features for segmentation without explicit semantic guidance or structural enhancement. On the COD10K dataset, this model achieves an $S_\alpha$ of 0.831, with limited boundary accuracy and object consistency. This indicates that in camouflaged object scenarios characterized by complex backgrounds and extremely weak saliency, relying solely on the visual backbone leads to attention drift and boundary blurring issues.

\subsubsection{Effect of CLIP (B $\rightarrow$ B+C)}
By incorporating the semantic mask $M_1$ generated by CLIP into the baseline (model b), the network is endowed with task-relevant linguistic priors during the feature extraction stage, thereby guiding attention to focus on potential camouflaged regions. This guidance effectively reduces interference from irrelevant background, leading to an improvement of $S_\alpha$ to 0.890, $E_{\phi}$ to 0.947, and an increase of 0.119 in $F^w_\beta$, indicating a significant enhancement in the model’s ability to capture the overall object structure.

\subsubsection{Effect of Fourier Edge Enhancement Module (B+C $\rightarrow$ B+C+E)}
By further introducing the FEEM (model c), Edge Enhancement features are explicitly extracted through multi-scale feature fusion and frequency-domain high-frequency enhancement. With FEEM incorporated, $F^w_\beta$ increases from 0.825 to 0.830, and $M$ decreases from 0.019 to 0.018, demonstrating the effectiveness of FEEM in enhancing boundary detail quality.

\subsubsection{Effect of Structure-Aware Attention Module and Coarse-Guided Local Refinement Module (B+C+E $\rightarrow$ B+C+E+SC)}
Finally, the SAAM and the CGLRM are introduced on top of model c (model d). Specifically, the SAAM module, guided by semantic and edge information, enhances cross-scale structural and boundary modeling, but using it alone still makes it difficult to ensure overall regional consistency. The CGLRM module generates coarse global guidance through global attention and performs local refinement via spatial partitioning, thereby improving boundary integrity and regional coherence. Since the two modules are complementary in design, they are introduced as a whole in the ablation study. Experimental results show that this combination improves $S_\alpha$ to 0.894 and $E_{\phi}$ to 0.950, and achieves the best performance across all metrics, further validating the effectiveness of this module combination.

\section{Conclusion}
This paper proposes a Language-Guided Structure-Aware Network, which integrates CLIP, the FEEM, the SAAM, and the CGLRM. The proposed framework effectively addresses the challenges of camouflaged object detection, including high similarity between objects and backgrounds, and missing local structures, achieving significant performance improvements on multiple benchmark datasets. However, the method still has certain limitations: the overall network structure is relatively complex, leading to high computational overhead. Future work may focus on designing more lightweight semantic guidance and boundary modeling mechanisms to further reduce inference costs and improve practical applicability.

% \section{CRediT authorship contribution statement}
% Min Zhang: Led study design; developed and implemented the algorithms; conducted model optimization.
% Yong Wang: Provided experimental resources; guided the study conception and design; offered overall supervision and critical manuscript revision.
% Boran Yang: analyzed data and results, and coordinated manuscript revision.
% Duoqian Miao: critically reviewed the manuscript.

\section{Declaration of competing interest}
All the authors declare that they have no competing financial interests or personal relationships that could influence the work reported in this paper.

\section{Data Availability}
The data for this study's findings are available online.

\section{Acknowledgements}
This work was supported by the National Natural Science Foundation of China (Grants 61976158 and 61673301) and the Chongqing Municipal Science and Technology Bureau, China (Grant No. CSTB2025TIAD-qykjggX0189).

\newpage
\printcredits

\bibliographystyle{unsrt}

\bibliography{menet}
\end{document}